\documentclass[11pt]{article}

\usepackage[T1]{fontenc}
\usepackage[utf8]{inputenc}
\usepackage{microtype}
\usepackage{booktabs}
\usepackage{amsmath,amssymb}
\usepackage{graphicx}
\usepackage[margin=1in]{geometry}
\usepackage{natbib}          
\usepackage[hidelinks]{hyperref}
\usepackage{url}

\newcommand{\claim}[1]{#1}   
\newcommand{\nodelta}{\textsuperscript{$\dagger$}}

\title{Structure, Association, and Decision Value:\\
Representation-Based Difficulty Estimation for\\
Adaptive Inference in African-Language NLI}

\author{
  Toheeb Ogunade\\
  University of Lagos\\
  \texttt{240805099@live.unilag.edu.ng}
}
\date{}

\begin{document}
\maketitle

\begin{abstract}
\claim{We ask whether internal representation statistics can provide useful
example-level difficulty signals for adaptive inference in multilingual African
NLP, and find that they cannot in this setting---for reasons that arise before
the representation analysis itself.} Studying natural language inference across
15 African languages with frozen off-the-shelf checkpoints, we report four
results. First, evaluation validity:
\claim{AfriXNLI's English configuration shares 1{,}047 of its 1{,}050 examples
verbatim with XNLI evaluation data, and one widely used NLI checkpoint scores
1.000 on that configuration's test split, a result consistent with XNLI test
exposure.} Because the benchmark is a translation of XNLI, its English, French
and Swahili configurations cannot serve as clean evaluations for XNLI-trained
models. Second, parameter count does not order capability across African
languages: our larger checkpoint is better in seven languages and worse in
eight, with no significant aggregate difference. Third, across three multilingual
representation spaces---a task-tuned model, an African-adapted masked language
model, and a generic one---angular dispersion is consistently dominated by
between-language variance while effective rank is not, so pooled correlations
inflate the former nearly fivefold and mask the latter. Fourth, the association
that survives language control is target-specific: effective rank predicts the
probability gain from escalation but not whether escalation changes the
prediction, while cheap-model confidence shows the opposite pattern; the two
formalisations of computational benefit correlate at only $0.655$. Under the
tested models, signals, and compute budgets, no evaluated signal makes adaptive
routing preferable to always-expensive inference, although an oracle exceeds it
by eleven accuracy points at 60\% of the compute.
\claim{Our central methodological finding is that a representation statistic can
be highly structured and statistically significant with respect to one notion of
computational benefit while being irrelevant to another, and therefore useless as
a decision variable.}
\end{abstract}

\section{Introduction}
\label{sec:intro}

Suppose a representation statistic computed from a model's hidden states
correlates with a plausible measure of example difficulty, across fifteen
languages and nine thousand examples, at a significance level that survives
multiple-comparison correction by thirty orders of magnitude. It is natural to
conclude that the statistic captures something about difficulty, and reasonable
to expect it to inform decisions that depend on difficulty---most obviously, how
much computation an input deserves.

This paper reports a case in which each step of that reasoning is defensible and
the conclusion is nonetheless wrong. The setting is adaptive inference for
low-resource multilingual natural language inference: given a cheap model and a
more expensive one, can the cheap model's internal representation tell us which
inputs are worth escalating? The question matters most in low-resource
multilingual deployment, where compute is scarce and per-language performance
varies widely, and it is precisely there that the assumptions behind such methods
are least often checked.

\claim{Before asking whether representation geometry predicts example
difficulty, one must establish that the evaluation and statistical setup can
distinguish example difficulty from language identity and from benchmark
contamination.} Our initial setup failed on both counts, and a third problem
emerged behind them; correcting each changed the answer. AfriXNLI, the benchmark we use,
shares \textbf{1{,}047 of its 1{,}050 English examples verbatim with XNLI}, and
the off-the-shelf checkpoints used to evaluate on it are trained on XNLI; its
English, French and Swahili configurations therefore cannot function as clean
evaluations, and one checkpoint answers every English test item correctly.
Pooled correlations between representation statistics and difficulty turned out
to be substantially carried by language identity. And our own routing experiment
optimised a predictor for one formalisation of ``benefit from computation'' while
scoring it against another.

\paragraph{This paper.}
We report a single investigation in four stages, each establishing why the
overall answer is negative:

\begin{enumerate}
  \item \textbf{Evaluation validity} (\S\ref{sec:f1}). AfriXNLI preserves XNLI
        examples verbatim, and contamination interacts with \emph{which} XNLI
        split each checkpoint was trained on, which the development--test
        accuracy gap exposes without access to training data.
  \item \textbf{Capability ordering} (\S\ref{sec:f2}). Parameter count does not
        reliably order accuracy across African languages, so the compute
        hierarchy a cascade presupposes cannot be assumed.
  \item \textbf{Language confounding and target dependence}
        (\S\ref{sec:f3a}--\S\ref{sec:f3b}). Across three multilingual
        representation spaces, some geometry statistics are dominated by language
        identity and others are not, so pooled analysis both manufactures and
        masks relationships; and the association that survives is specific to
        which notion of computational benefit it is measured against.
  \item \textbf{Adaptive-routing utility} (\S\ref{sec:f4}). No signal we tested
        makes routing preferable to always-expensive inference, though an oracle
        shows the allocation problem is learnable.
\end{enumerate}

\paragraph{What the reader should take away.}
\claim{A representation statistic can be highly structured and statistically
significant without being a useful decision variable.} \claim{Which notion of
``benefit from computation'' one measures determines which signals appear
informative.}

\section{Background and Related Work}
\label{sec:related}

\paragraph{African-language NLI.} Evaluation resources for African languages have
expanded substantially, from named-entity benchmarks
\citep{adelani-etal-2021-masakhaner} to broader multi-task suites. IrokoBench
\citep{adelani-etal-2025-irokobench} introduces AfriXNLI, the natural language inference
benchmark used here, by translating a subset of XNLI
\citep{conneau-etal-2018-xnli} into 16 African languages. On the modelling side,
XLM-R \citep{conneau-etal-2020-unsupervised} provides the multilingual encoder that most
African-adapted models extend, including AfroXLMR
\citep{alabi-etal-2022-adapting}, which adapts it through continued pre-training on
African-language corpora. The translated construction that makes AfriXNLI
possible is also what creates the lineage we examine in \S\ref{sec:f1}.

\paragraph{Multilingual representation geometry.} A line of work characterises
contextual representation spaces through geometric summaries---anisotropy,
directional concentration, and spectral or rank-based measures of how many
dimensions a representation effectively occupies
\citep{ethayarajh-2019-contextual}. Such statistics are attractive as difficulty or
quality proxies because they are cheap to compute from a forward pass that has
already been performed. They are also, in multilingual corpora, routinely
reported pooled across languages, which is the practice \S\ref{sec:f3a}
examines.

\paragraph{Adaptive inference and cascades.} Conditional computation for
transformer inference is well established, and we claim no novelty in the idea.
Early-exit methods attach classifiers to intermediate layers and halt when a
confidence criterion is met \citep{xin-etal-2020-deebert} or when successive layers
agree \citep{zhou2020pabee}. CascadeBERT \citep{li-etal-2021-cascadebert-accelerating} argues that
intermediate-layer representations are unreliable for this purpose and instead
cascades complete models, calibrating them so that a cheap model's confidence can
govern escalation---the design our compute ladder follows. Calibration itself
follows standard temperature scaling \citep{guo2017calibration}. Our contribution
is not a new routing mechanism but an examination of whether representation
statistics supply a usable signal for one.

\paragraph{Benchmark contamination.} Overlap between evaluation data and training
corpora is a recognised threat to benchmark validity, with recent work arguing
that contamination should be measured and reported per benchmark rather than
assumed absent \citep{sainz2023contamination}, and proposing procedures for
detecting it \citep{golchin2023quiz}. Translated benchmarks are an underexamined
case: contamination can propagate through the source examples even when no target
string was ever seen. Separately, the distinction between pooled and
within-group association is long-standing outside NLP---the ecological fallacy
\citep{robinson1950ecological} names precisely the inference we find multilingual
representation analyses at risk of making.

\section{Experimental Design}
\label{sec:design}

\subsection{Data and the clean-language protocol}
AfriXNLI \citep{adelani-etal-2025-irokobench} provides 18 language configurations with 450
development and 600 test examples each and perfectly balanced three-way labels.
XNLI intersects AfriXNLI at \{\texttt{eng}, \texttt{fra}, \texttt{swa}\}; these
are excluded throughout. The remaining 15 configurations (\texttt{amh},
\texttt{ewe}, \texttt{hau}, \texttt{ibo}, \texttt{kin}, \texttt{lin},
\texttt{lug}, \texttt{orm}, \texttt{sna}, \texttt{sot}, \texttt{twi},
\texttt{wol}, \texttt{xho}, \texttt{yor}, \texttt{zul}) form the clean set:
$n=6{,}750$ development and $n=9{,}000$ test examples.

\subsection{Models}
Table~\ref{tab:models} lists every checkpoint used. Three serve as rungs of the
compute ladder and three as sources of representation statistics; mDeBERTa-base
appears in both roles, as a rung and as a feature source. No checkpoint was
fine-tuned, and the feature sources are never asked to produce a prediction.

\begin{table}[t]
\centering
\small
\begin{tabular}{llrrr}
\toprule
Role & Checkpoint & Layers & Hidden & Params \\
\midrule
Cheap rung   & multilingual-MiniLMv2-L6-mnli-xnli & 6  & 384  & $\sim$118M \\
Mid rung     & mDeBERTa-v3-base-xnli-\dots-2mil7  & 12 & 768  & $\sim$278M \\
Large rung   & xlm-roberta-large-xnli             & 24 & 1024 & $\sim$560M \\
\midrule
Features (task-tuned)  & mDeBERTa-v3-base & 12 & 768 & --- \\
Features (African MLM) & afro-xlmr-base   & 12 & 768 & --- \\
Features (generic MLM) & xlm-roberta-base & 12 & 768 & --- \\
\bottomrule
\end{tabular}
\caption{Frozen checkpoints \citep{wang-etal-2021-minilmv2,he2023debertav3,
conneau-etal-2020-unsupervised,alabi-etal-2022-adapting}; no fine-tuning was
performed. The three feature
sources supply representation statistics only and are never asked to predict.
\texttt{xlm-roberta-base} was added post hoc as a replication source
(\S\ref{sec:f3a}).}
\label{tab:models}
\end{table}

\subsection{Label alignment and calibration}
Checkpoint label orders differ: \texttt{mDeBERTa} and \texttt{MiniLM} emit
$(\text{entailment}, \text{neutral}, \text{contradiction})$ while
\texttt{xlm-roberta-large-xnli} emits the reverse. All logits are permuted into
the AfriXNLI canonical order before any comparison. We note this because it fails
silently: comparing probabilities by logit index across these checkpoints
subtracts one class from a different one and yields plausible-looking quantities
with no meaning.

Each model receives one scalar temperature fitted by L-BFGS on the pooled clean
\emph{development} split ($n=6{,}750$): $T=4.116$ (MiniLM), $T=1.704$
(mDeBERTa), $T=2.723$ (XLM-R-large). All reported results are on test.

\subsection{The escalation target}
\label{sec:delta}
For cheap model $s$ and expensive model $e$ we measure the marginal value of
escalation as the change in the calibrated probability assigned to the gold
label $y^\star$:
\begin{equation}
\Delta_{\mathrm{prob}}(x) \;=\; p_e(y^\star \mid x) \;-\; p_s(y^\star \mid x).
\label{eq:dprob}
\end{equation}
We use a continuous quantity rather than a change in correctness \emph{for
measuring association}, because the latter takes only three values and is
dominated by examples near the decision boundary, which costs a great deal of
statistical power. That choice is appropriate for the correlation analysis and,
as \S\ref{sec:twotargets} and \S\ref{sec:targetalign} show, inappropriate for
training a router: the two are different quantities, and conflating them was an
error in our initial design.

\subsection{Which results depend on a constructed target}
\label{sec:dependency}
Both targets used in this paper---$\Delta_{\mathrm{prob}}$ above and
$\Delta_{\mathrm{correct}}$, introduced in \S\ref{sec:twotargets}---are
constructed from a single frozen checkpoint pair rather than the seed-averaged
quantity the design called for (\S\ref{sec:limitations}). We therefore mark
explicitly which results depend on them.

\begin{table}[t]
\centering
\small
\begin{tabular}{llc}
\toprule
\S & Result & Depends on a target \\
\midrule
\ref{sec:f1}  & Benchmark contamination        & no \\
\ref{sec:f2}  & Capability not ordered by size & no \\
\ref{sec:f3a} & Language dominates geometry ($\eta^2$) & no \\
\ref{sec:f3b} & Residual association with either target & yes \\
\ref{sec:f4}  & Routing evaluation             & yes \\
\bottomrule
\end{tabular}
\caption{Sections marked ``no'' are computed from accuracies, string matching,
and representations alone. A reader who rejects our target construction may
discard \S\ref{sec:f3b}--\S\ref{sec:f4} without affecting the rest.}
\label{tab:dependency}
\end{table}

\subsection{Representation statistics}
For each example, hidden states at layers $\{4,8,12\}$ over non-padding tokens
give $X \in \mathbb{R}^{T \times d}$. With $p_i$ the normalised squared singular
values of column-centred $X$, and $U$ the row-normalised $X$:
\begin{align*}
\text{effective rank} &= \exp\!\Big(-\textstyle\sum_i p_i \log p_i\Big),\\
\text{spectral concentration} &= p_1,\\
\text{angular dispersion} &= 1 - \big\lVert \tfrac{1}{T}\textstyle\sum_t U_t \big\rVert.
\end{align*}
Three sources $\times$ three layers $\times$ three statistics gives 27 features.

\subsection{Statistical procedure}
Associations between a representation statistic and a target are measured by
partial Spearman correlation, controlling for cheap-model confidence, token count
and subword fragmentation by regressing all quantities on the ranks of the
controls and correlating the residuals. Every such correlation is computed
\emph{within} each language and the 15 per-language estimates are then combined by
Fisher-$z$ meta-analysis weighted by $n - k - 3$, yielding a pooled within-language
estimate with a confidence interval and a two-sided $p$-value. Because 27 features
are screened against each target, all $p$-values are Holm-corrected across
features. Between-language structure is quantified by $\eta^2$
(Equation~\ref{eq:eta2}), the share of a quantity's total variance lying between
languages.

Uncertainty is estimated by resampling rather than by parametric assumption.
Correlation stability uses a within-language bootstrap and subsampling at 25\% and
50\%; a label-permutation null establishes the scale of spurious association.
Accuracy differences between models use a cluster bootstrap that resamples
languages and then examples within language, since examples within a language are
not independent and the comparison generalises over languages. Routing is
evaluated leave-one-language-out at matched average compute budget: the predictor
is fitted on 10 languages and applied to the held-out one, and every method is
compared at the same escalation rate.

\subsection{Benchmark lineage and contamination audit}
\label{sec:audit}
Before any modelling result, we audit the benchmark itself in three steps.
First, exact-string matching of every AfriXNLI premise--hypothesis pair in the
overlapping configurations against the XNLI evaluation splits, which establishes
lineage independently of any documentation. Second, cross-referencing the
training-data statements on each checkpoint's model card against the splits that
matching implicates. Third, split-specific accuracy probing: measuring each model
separately on the development and test configurations, since a model exposed to
one split but not the other should show a gap that a genuinely capable model would
not. The three steps are cheap, require no access to training corpora, and in our
case each was necessary---the lineage alone does not identify which models are
affected, and the cards alone do not identify which splits. We suggest this
sequence should precede evaluation on any translated benchmark.

\section{Stage I: Evaluation Validity}\nodelta
\label{sec:f1}
{\renewcommand{\thefootnote}{}\footnotetext{$\dagger$ Results marked with this
symbol do not depend on the construction of either marginal-gain target
($\Delta_{\mathrm{prob}}$, $\Delta_{\mathrm{correct}}$); see
\S\ref{sec:dependency}.}\addtocounter{footnote}{-1}}

\claim{AfriXNLI is a translation of XNLI, and XNLI-trained checkpoints therefore
cannot be evaluated cleanly on its English, French, or Swahili configurations.}

An evaluation can only support conclusions about model behaviour to the extent
that it measures behaviour rather than recall. We therefore begin by establishing
what our evaluation measures, before using it to interpret anything about
representations.

\subsection{Benchmark lineage}

AfriXNLI is constructed by translating a subset of XNLI into 16 African
languages, and its own documentation states that it retains the original English
and French subsets and that its \texttt{dev} and \texttt{test} splits are subsets
of the corresponding XNLI splits. We confirm this by exact string matching: of
the 1{,}050 unique English premise--hypothesis pairs in AfriXNLI, \textbf{1{,}047
(99.7\%)} occur verbatim in the XNLI evaluation splits, and the 450-example
AfriXNLI English development split is \emph{exactly} XNLI validation. Because
XNLI covers 15 languages including Swahili and French, the two benchmarks
intersect at \{\texttt{eng}, \texttt{fra}, \texttt{swa}\}.

This lineage is public and, in itself, unsurprising: translated benchmarks are
built from source benchmarks. It becomes consequential when combined with the
training data of the checkpoints used to evaluate on it. The model cards for both
cheap rungs state training on XNLI---MiniLM specifically on ``the XNLI
development dataset and the MNLI train dataset''---and the large rung's card
states fine-tuning on ``a combination of NLI data in 15 languages'' without
identifying which splits.

\subsection{Split-specific contamination}

\claim{The contamination is split-specific, and the gap between development and
test accuracy acts as a fingerprint of which split a model was exposed to.}

\begin{table}[t]
\centering
\small
\begin{tabular}{lcccccc}
\toprule
& \multicolumn{2}{c}{\texttt{eng}} & \multicolumn{2}{c}{\texttt{fra}}
& \multicolumn{2}{c}{\texttt{swa}} \\
\cmidrule(lr){2-3}\cmidrule(lr){4-5}\cmidrule(lr){6-7}
Model & dev & test & dev & test & dev & test \\
\midrule
MiniLM-L6     & 0.887 & 0.760 & 0.927 & 0.720 & 0.891 & 0.622 \\
mDeBERTa-base & 1.000 & 0.890 & 0.998 & 0.867 & 0.993 & 0.742 \\
XLM-R-large   & 0.998 & \textbf{1.000} & 0.987 & \textbf{0.995} & 0.973 & \textbf{0.978} \\
\bottomrule
\end{tabular}
\caption{Accuracy on the contaminated configurations, full splits
($n=450$ dev, $n=600$ test). The two models whose cards report training on XNLI
\emph{development} data lose 11--27 points from dev to test. XLM-R-large loses
nothing.}
\label{tab:devtest}
\end{table}

Table~\ref{tab:devtest} separates the three checkpoints sharply. MiniLM and
mDeBERTa perform far better on the development configurations than on the test
configurations---mDeBERTa is perfect on English dev at $1.000$ and falls to
$0.890$ on English test, and the corresponding Swahili drop is 25 points. This is
the pattern their cards predict: both report training on XNLI development data,
and AfriXNLI development \emph{is} XNLI validation.

XLM-R-large shows no such gap. It scores $0.998$ on English dev and $1.000$ on
English test, $0.973$ and $0.978$ on Swahili. \claim{This is consistent with
exposure to XNLI test data as well as development data}, but we state it as a
consistency rather than a fact: the model card does not specify which splits were
used, and we cannot verify training data we do not have access to. What can be
said without inference is that the checkpoint answers every English test item
correctly, and that no model in our study achieves anything remotely comparable
on uncontaminated languages, where the same checkpoint reaches $0.523$.

\subsection{What follows, and what does not}

Three claims of decreasing certainty are involved here, and we separate them
deliberately. The \emph{lineage}---that AfriXNLI is derived from XNLI and shares
its evaluation examples---is both documented by the dataset and measured by us
through exact string matching; it is established. The \emph{development-split
exposure} of MiniLM and mDeBERTa is stated on their model cards and independently
corroborated by the dev--test gap in Table~\ref{tab:devtest}; it is likewise
established. The \emph{test-split exposure} of XLM-R-large is neither documented
nor directly verifiable, and is inferred from behaviour alone; it is not
established, and nothing in the remainder of this paper depends on it. Readers
who reject the third claim should still treat the first two as grounds for
excluding the affected configurations, since exclusion follows from the lineage
regardless of any individual model's training history.

The practical consequence is that \texttt{eng}, \texttt{fra}, and \texttt{swa}
cannot serve as clean evaluations---or, more importantly for this study, as the
high-resource control against which African-language behaviour is compared. We
exclude them from all subsequent analysis and work with the 15 remaining
configurations.

That exclusion does not make the remainder pristine.
\claim{For the 15 African configurations this is benchmark-lineage contamination
rather than verbatim leakage}: the surface strings are translations the models
have not seen, but the underlying premise--hypothesis pairs were seen in other
languages. A model with strong cross-lingual transfer may therefore retain some
advantage on them. We report this as a limitation rather than a controlled
variable, since no uncontaminated African-language NLI benchmark of comparable
coverage was available to us.

Having established which portion of the benchmark can support interpretation,
we next ask whether the models themselves furnish the compute hierarchy that an
adaptive-inference study presupposes.

\section{Stage II: Capability Is Not Ordered by Parameter Count}\nodelta
\label{sec:f2}

\claim{Among off-the-shelf XNLI checkpoints, model size does not predict which
model wins on a given African language; the ordering is language-dependent and
unstable.}

An adaptive-inference system presupposes a hierarchy of computational
alternatives: escalation must purchase something. The most convenient way to
construct such a hierarchy is by parameter count, treating a larger checkpoint as
the more capable one. This section asks whether that assumption holds in the
setting we are studying.

\subsection{Parameter count is not a sufficient proxy}

\begin{table}[t]
\centering
\small
\begin{tabular}{lrrrr}
\toprule
Model & Params & Layers & Accuracy & Above chance \\
\midrule
MiniLMv2-L6      & $\sim$118M & 6  & 0.410 & 11/15 \\
mDeBERTa-v3-base & $\sim$278M & 12 & 0.545 & 14/15 \\
XLM-R-large      & $\sim$560M & 24 & 0.523 & 14/15 \\
\bottomrule
\end{tabular}
\caption{Accuracy on the 15 clean languages ($n=9{,}000$). Chance is $0.333$;
``above chance'' counts languages whose Wilson 95\% lower bound exceeds it.}
\label{tab:capability}
\end{table}

Table~\ref{tab:capability} shows that accuracy is not monotone in size. The
largest checkpoint, with roughly twice the parameters and twice the depth of
mDeBERTa-base, is the second-best of the three. We do not claim it is worse. The
aggregate difference of $+0.022$ in mDeBERTa's favour is not statistically
distinguishable from zero: a cluster bootstrap resampling languages and then
examples within language (5{,}000 replicates) gives a 95\% confidence interval of
$[-0.035, +0.083]$, with $P(\mathrm{diff} > 0) = 0.772$.\footnote{An
example-level bootstrap gives $[+0.009, +0.035]$, apparently excluding zero. That
interval is wrong for this comparison: examples within a language are not
independent draws, and the quantity of interest generalises over languages rather
than over items within the 15 we happen to have. We report it only to note that
the choice of resampling unit changes the conclusion.} The defensible statement
is weaker and more useful than an ordering: parameter count simply does not
determine which of these two models is better.

\subsection{The better model is language-dependent}

\claim{A model can be substantially better than another for one language and
substantially worse for a second.} Per-language accuracy differences between
mDeBERTa and XLM-R-large span
$-0.145 \le \Delta_{\mathrm{acc}} \le +0.220$
(Table~\ref{tab:perlang}, Appendix~\ref{app:perlang}): XLM-R-large is 14.5 points better on Amharic and
12.7 on Oromo, while mDeBERTa is 22.0 points better on Igbo and 20.0 on Shona.
Seven languages favour mDeBERTa and eight favour XLM-R-large, and in six of the
fifteen the difference exceeds ten accuracy points in one direction or the other.
The aggregate near-tie is thus the average of large and inconsistent
disagreements rather than the average of small ones.

We have not identified which linguistic, script, or resource properties predict
the direction of the difference, and with 15 languages and three checkpoints we are
not positioned to: the observation here is that the ordering varies, not why.

\subsection{Consequences for adaptive inference}

This matters directly for the framing of the present study.
\claim{If the expensive model is not reliably more capable, escalation cannot be
interpreted as purchasing additional intelligence; it purchases a different model
whose value depends on the language and input regime.} A cascade ordered by
parameter count is, in this setting, not a compute ladder but a model switch with
a language-dependent sign.

The consequence is measurable. For the mDeBERTa$\rightarrow$XLM-R-large pair,
$\Delta_{\mathrm{prob}}$ has mean $-0.011$: escalation helps on 37.5\% of
examples and hurts on 39.7\%, which is not a target a router can profitably
predict because there is no aggregate benefit to allocate. Only the
MiniLM$\rightarrow$mDeBERTa pair forms a usable ladder---approximately
$8\times$ the floating-point operations, mean $\Delta_{\mathrm{prob}} = +0.099$
(sd $0.221$), helping on 51.2\% of examples and hurting on 26.0\%---and even that
holds only for the 11 of 15 languages in which the cheap rung exceeds chance. All
routing experiments in \S\ref{sec:f4} are therefore restricted to those 11
languages, which is a real limitation on their generality.

We emphasise that this section establishes nothing about routing itself. It
establishes that the compute hierarchy an adaptive-inference study presupposes
cannot be assumed from model size and must be verified per language before any
routing claim is attempted.

If model capacity cannot be ordered reliably by parameter count, and if
representation statistics may themselves carry language-level structure, then the
next question is prior to any difficulty analysis: are the internal variables such
analyses rely on primarily properties of examples, or of the languages those
examples are expressed in?

\section{Stage III(a): Language Identity Dominates Representation Geometry}\nodelta
\label{sec:f3a}

A representation statistic intended to measure something about an \emph{example}
should vary primarily with the example. In a multilingual corpus it may instead
vary primarily with the \emph{language} the example is written in, in which case
any quantity computed by pooling examples across languages will partly be a
restatement of language identity. This stage measures how much of each
statistic's variation is attributable to language, before any difficulty target
is introduced. It requires only representations and language labels, and is
therefore independent of the modelling choices examined in
\S\ref{sec:f3b}--\S\ref{sec:f4}.

For each statistic we compute $\eta^2$, the proportion of its total variance
lying between languages rather than within them:
\begin{equation}
\eta^2 \;=\; \frac{\sum_{\ell} n_\ell (\bar{v}_\ell - \bar{v})^2}
                  {\sum_{i} (v_i - \bar{v})^2},
\label{eq:eta2}
\end{equation}
where $\ell$ indexes the 15 clean languages, $n_\ell$ is the number of test
examples in language $\ell$, and $\bar{v}_\ell$ is the mean of the statistic
within that language. A value near zero indicates a statistic that varies almost
entirely example-to-example; a value near one indicates one that is close to a
per-language constant.

\claim{Across three multilingual representation spaces, angular dispersion is
consistently the most language-determined statistic and effective rank
consistently the least.}

\begin{table}[t]
\centering
\small
\begin{tabular}{lccc}
\toprule
& mDeBERTa-base & AfroXLMR-base & XLM-R-base \\
Statistic & (task-tuned) & (African MLM) & (generic MLM) \\
\midrule
Angular dispersion     & \textbf{0.583} & \textbf{0.484} & \textbf{0.542} \\
Spectral concentration & 0.112 & 0.249 & 0.241 \\
Effective rank         & 0.075 & 0.133 & 0.151 \\
\bottomrule
\end{tabular}
\caption{Between-language variance share $\eta^2$, averaged over layers
$\{4,8,12\}$. The ordering replicates in all three representation spaces: a
task-tuned model, an African-adapted MLM, and a generic multilingual MLM.}
\label{tab:eta2}
\end{table}

\begin{figure}[t]
\centering
\includegraphics[width=0.85\textwidth]{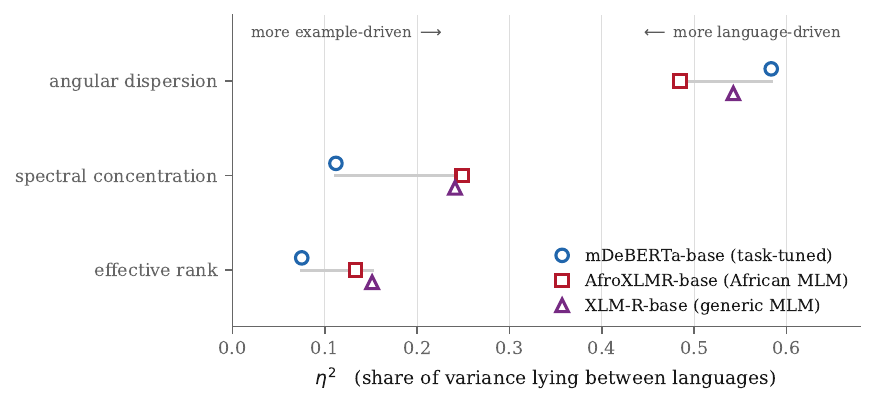}
\caption{Between-language variance share $\eta^2$ for each geometry statistic in
each representation space, averaged over layers $\{4,8,12\}$; the values are
those of Table~\ref{tab:eta2}. The ordering---angular dispersion most
language-determined, effective rank least---is identical in all three spaces,
although the magnitudes differ.}
\label{fig:eta2}
\end{figure}

Table~\ref{tab:eta2} and Figure~\ref{fig:eta2} report $\eta^2$ averaged over
layers $\{4,8,12\}$ for each statistic in each representation space. The three sources were chosen to differ
in how they were produced: \texttt{mDeBERTa-v3-base} is fine-tuned for NLI,
\texttt{afro-xlmr-base} is a masked language model adapted to African languages,
and \texttt{xlm-roberta-base} is a generic multilingual masked language model
with no African adaptation and no task supervision. They share no training
objective, and only the last two share a pre-training corpus.

The magnitudes differ across sources, but the ordering does not. Angular
dispersion is the most language-determined statistic in all three spaces
($\eta^2 = 0.583$, $0.484$, $0.542$), and effective rank the least
($0.075$, $0.133$, $0.151$), with spectral concentration intermediate
throughout. At the individual-layer level the effect is more pronounced still:
angular dispersion at layer 8 of mDeBERTa reaches $\eta^2 = 0.757$.

The implication is direct. Roughly half of the variation in angular dispersion,
and three quarters of it at some layers, is a property of the language rather
than of the example. A quantity computed by pooling such a statistic across
languages is therefore substantially a re-description of language identity, and
will behave like a difficulty signal to the exact extent that languages differ
in difficulty. This holds whatever target the statistic is subsequently
correlated against, which is why the observation does not depend on our choice
of target: it is a property of the representations, not of the analysis
downstream of them.

We emphasise what is and is not established here. $\eta^2$ is descriptive; it
identifies how variance is distributed, not why. We do not claim that angular
dispersion \emph{encodes} language identity in any mechanistic sense, nor that
it is uninformative about examples---only that its example-level variation is a
minority of its total variation, and that pooled analyses cannot separate the
two. Nor do the three sources constitute an exhaustive sample: they are three
encoder models of similar depth, and the pattern may not extend to decoder-only
or substantially larger models.

\paragraph{A stronger claim that did not replicate.} Having observed this
pattern on two sources, we hypothesised a quantitative version of it: that
$\eta^2$ predicts, across features, how far a pooled correlation departs from
the corresponding within-language one. We added the third source specifically to
test that hypothesis, and it did not survive.

Across all 27 features,
$\mathrm{Spearman}(\eta^2, |\rho_{\text{pooled}}|-|\rho_{\text{within}}|)=+0.415$
with a permutation $p$ of $0.032$, which taken alone would appear to support the
hypothesis. The bootstrap 95\% confidence interval, however, is
$[-0.031, +0.737]$, and the relationship is carried almost entirely by a single
source: $+0.917$ for mDeBERTa against $+0.133$ for both AfroXLMR and
XLM-R-base. The two procedures disagree because the 27 features are not
independent---nine per source, sharing layers and statistics---which the
permutation test over feature labels does not accommodate. \claim{We therefore
do not claim that $\eta^2$ predicts pooling bias as a general quantitative
law.}

We report this failure because it delimits the surviving claim. What replicates
across three representation spaces is an ordering of statistics by their
language dependence, not a quantitative relationship between that dependence and
the resulting analytical bias. The former is sufficient for the practical
recommendation we draw in \S\ref{sec:discussion}; the latter would have been a
stronger and more general result, and we did not obtain it.

\section{Stage III(b): Representation Signals Depend on What
``Benefit from Computation'' Means}
\label{sec:f3b}

From this point the analysis depends on a constructed target, and therefore on
the modelling choices discussed in \S\ref{sec:delta} and
\S\ref{sec:limitations}. The results in \S\ref{sec:f1}--\S\ref{sec:f3a} do not.

\subsection{Two notions of marginal benefit}
\label{sec:twotargets}

An input that ``benefits from more computation'' can be characterised in at least
two ways. The first asks how much escalation moves the model's probability on the
correct label; the second asks whether escalation changes the decision at all:
\begin{align}
\Delta_{\mathrm{prob}}(x) &= p_e(y^\star \mid x) - p_s(y^\star \mid x),
  \tag{\ref{eq:dprob}, restated}\\
\Delta_{\mathrm{correct}}(x) &= \mathbf{1}[\hat{y}_e = y^\star]
                              - \mathbf{1}[\hat{y}_s = y^\star].
  \label{eq:dcorr}
\end{align}
Equation~\ref{eq:dprob} is continuous (mean $+0.099$, sd $0.221$ on the viable
ladder); Equation~\ref{eq:dcorr} is ternary, taking $+1$ on 24.7\% of examples,
$-1$ on 11.2\%, and $0$ on the remaining 64.1\%. We refer to
$\Delta_{\mathrm{correct}}$ as the \emph{decision-relevant} or \emph{routing}
target, because it is the quantity a routing system's accuracy depends on. We do
not call it the true target: both are legitimate operationalisations of
computational benefit, and part of what follows is that the choice matters.

The two are correlated at $+0.655$---substantial, but far from interchangeable.
They also differ in how much of their variation is attributable to language:
$\eta^2 = 0.126$ for $\Delta_{\mathrm{prob}}$ against $0.033$ for
$\Delta_{\mathrm{correct}}$, so the choice of target additionally determines how
much of the language confounding established in \S\ref{sec:f3a} is inherited.

\subsection{Pooled analysis is inadequate}

\claim{Naive pooled analysis errs in both directions: it manufactures a
relationship for the language-determined statistic and erases a genuine one for
the language-neutral statistic.}

\begin{table}[t]
\centering
\small
\begin{tabular}{lccc}
\toprule
Feature (mDeBERTa) & $\eta^2$ & Pooled $\rho$ & Within $\rho$ \\
\midrule
L8 angular dispersion      & 0.757 & $+0.296$ & $+0.061$ \\
L12 effective rank         & 0.040 & $-0.009$ & $-0.127$ \\
L12 spectral concentration & 0.031 & $+0.044$ & $+0.099$ \\
\bottomrule
\end{tabular}
\caption{Partial Spearman with $\Delta_{\mathrm{prob}}$, controlling for
cheap-model confidence, token count, and fragmentation. Angular dispersion is
inflated $4.9\times$ by pooling; effective rank is masked roughly tenfold.}
\label{tab:simpson}
\end{table}

\begin{figure}[t]
\centering
\includegraphics[width=0.85\textwidth]{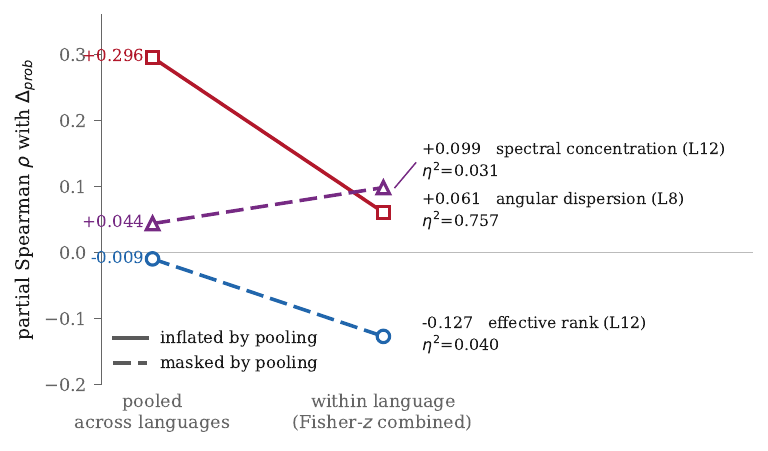}
\caption{Pooled versus within-language partial Spearman correlation with
$\Delta_{\mathrm{prob}}$, for the statistics of Table~\ref{tab:simpson}. Solid
lines mark associations inflated by pooling, dashed lines those masked by it.
Angular dispersion, the most language-determined statistic
($\eta^2 = 0.757$), loses most of its apparent relationship once language is
controlled; effective rank, the least ($\eta^2 = 0.040$), acquires one.}
\label{fig:pooled}
\end{figure}

Table~\ref{tab:simpson} and Figure~\ref{fig:pooled} show the consequence of
\S\ref{sec:f3a} for correlation analysis. Angular dispersion, the statistic that is 76\% between-language
variance at this layer, carries a pooled correlation of $+0.296$ with
$\Delta_{\mathrm{prob}}$ that falls to $+0.061$ once language is controlled.
Effective rank, at $\eta^2 = 0.040$, shows the opposite: a pooled correlation
indistinguishable from zero ($-0.009$) that becomes $-0.127$ under the same
control. An analyst screening candidate features by pooled correlation would promote
angular dispersion and discard effective rank, which is the opposite of what the
within-language evidence supports. The inversion is not confined to
$\Delta_{\mathrm{prob}}$: for effective rank against $\Delta_{\mathrm{correct}}$
the pooled correlation is $+0.050$ while the within-language estimate is
$-0.027$, a sign reversal.

\subsection{The surviving association is target-specific}
\label{sec:swap}

\claim{Effective rank predicts the probability gain from escalation but not
whether escalation changes the prediction; confidence shows the reverse
pattern.}

\begin{table}[t]
\centering
\small
\begin{tabular}{lcccc}
\toprule
& \multicolumn{2}{c}{$\Delta_{\mathrm{prob}}$}
& \multicolumn{2}{c}{$\Delta_{\mathrm{correct}}$} \\
\cmidrule(lr){2-3}\cmidrule(lr){4-5}
Signal & $\rho$ & Holm $p$ & $\rho$ & Holm $p$ \\
\midrule
mDeBERTa L12 effective rank      & $\mathbf{-0.127}$ & $5.8\!\times\!10^{-32}$ & $-0.027$ & $0.28$ \\
mDeBERTa L12 spectral conc.      & $+0.099$ & $2.9\!\times\!10^{-19}$ & $+0.003$ & $1.00$ \\
mDeBERTa L8 angular dispersion   & $+0.061$ & $2.2\!\times\!10^{-7}$  & $+0.007$ & $1.00$ \\
mDeBERTa L12 angular dispersion  & $+0.029$ & $0.13$ & $\mathbf{-0.066}$ & $1.6\!\times\!10^{-8}$ \\
\midrule
Cheap-model confidence           & $+0.003$ & --- & $\mathbf{-0.099}$ & --- \\
\bottomrule
\end{tabular}
\caption{Within-language partial Spearman against both targets, identical
procedure, Holm-corrected across all 27 features. Bold marks associations
significant at the Holm-adjusted level. The two targets are predicted by
different signals.}
\label{tab:swap}
\end{table}

The magnitudes in Table~\ref{tab:swap} are less interesting than the pattern.
Effective rank at layer 12 has the strongest association in the study with
$\Delta_{\mathrm{prob}}$ ($\rho = -0.127$, 95\% CI $[-0.147, -0.106]$,
Holm-adjusted $p = 5.8 \times 10^{-32}$, sign-consistent in 12/15 languages) and
survives bootstrap resampling, subsampling to a quarter of the data,
leave-one-language-out, and label permutation
(Table~\ref{tab:stability}, Appendix~\ref{app:stability}). Against $\Delta_{\mathrm{correct}}$ the same
feature reaches only $\rho = -0.027$ and does not survive multiple-comparison
correction ($p = 0.28$).

Cheap-model confidence exhibits the mirror image. Within language it is
unrelated to $\Delta_{\mathrm{prob}}$ ($\rho = +0.003$, CI $[-0.018, +0.023]$)
while being the strongest single predictor of $\Delta_{\mathrm{correct}}$
($\rho = -0.099$, CI $[-0.120, -0.079]$). A signal that appears uninformative
under one operationalisation of computational benefit is the best available
signal under the other.

Angular dispersion at layer 12 makes the point harder to dismiss as a magnitude
artifact: it is not significant against $\Delta_{\mathrm{prob}}$
($\rho = +0.029$, $p = 0.13$) but is strongly significant against
$\Delta_{\mathrm{correct}}$ with the \emph{opposite sign}
($\rho = -0.066$, $p = 1.6 \times 10^{-8}$, sign-consistent in 12/15 languages).
\claim{A representation statistic can predict one notion of computational
benefit while being irrelevant to---or oppositely related to---another.}

\paragraph{Registration.} \claim{This dual-target analysis was added after the
routing-target mismatch described in \S\ref{sec:targetalign} was identified, and
should be interpreted as a post-hoc analysis rather than a pre-specified test.}
It rests on a single cheap--expensive pair and has not been replicated across
model pairs.

We therefore asked not whether effective rank is statistically associated with
a marginal-benefit target---it is---but whether any of these associations carries
enough incremental information to support a useful routing decision.

\section{Stage IV: From Statistical Association to Adaptive Routing}
\label{sec:f4}

\subsection{Target alignment}
\label{sec:targetalign}

In our initial routing experiment we trained the routing predictor against
$\Delta_{\mathrm{prob}}$, the change in gold-label probability, but evaluated
routing quality using prediction correctness. We subsequently recognised that
these are different objectives. The predictor was therefore optimised for a
quantity that the evaluation did not measure.

The distinction is empirical rather than semantic. As reported in
\S\ref{sec:twotargets}, the two targets correlate at only $+0.655$, and the
dual-target analysis in \S\ref{sec:swap}---which we ran precisely because this
mismatch came to light---shows that they are predicted by different signals.
Cheap-model confidence is essentially unrelated to $\Delta_{\mathrm{prob}}$
within language ($\rho = +0.003$) while being the strongest available predictor
of $\Delta_{\mathrm{correct}}$ ($\rho = -0.099$). Effective rank shows the
opposite profile.

This mismatch gave the confidence baseline an advantage that representation-based
routing did not receive: confidence was related to the decision-relevant
correctness flip that the evaluation scored, while the geometry predictor had
been fitted to a quantity that is not. We therefore repeated the routing
experiment using $\Delta_{\mathrm{correct}}$ as the training target, holding
everything else fixed.

\claim{A routing method should be trained against the quantity whose decision
value is being evaluated.} We state this as a principle because our original
design violated it, and because the violation was not visible from the routing
results alone---it surfaced only when the two targets were analysed side by side.
We report the sequence in full---original hypothesis, initial result, recognition
of the objective mismatch, corrected target, re-evaluated conclusion---because
the correction is part of the evidence rather than an erratum.

\claim{The original comparison therefore does not support the stronger claim that
confidence is intrinsically superior to representation geometry. The appropriate
conclusion is that target alignment materially changes the comparison}, which
motivates the corrected experiment below.

\subsection{The corrected comparison}
\label{sec:corrected}

\begin{table}[t]
\centering
\small
\begin{tabular}{lrrrrrr}
\toprule
Budget & Random & Conf. & Geom.$(\Delta_{\mathrm{prob}})$
& Geom.$(\Delta_{\mathrm{correct}})$ & Geom.+Conf. & Oracle \\
\midrule
20\% & 0.457 & 0.466 & 0.461 & 0.468 & \textbf{0.472} & 0.613 \\
40\% & 0.486 & 0.506 & 0.498 & 0.504 & \textbf{0.508} & 0.688 \\
60\% & 0.519 & 0.543 & 0.524 & 0.535 & \textbf{0.545} & 0.688 \\
80\% & 0.548 & 0.564 & 0.553 & 0.559 & \textbf{0.565} & 0.688 \\
\bottomrule
\end{tabular}
\caption{Leave-one-language-out routing accuracy at matched compute budget,
11 viable ladder languages. Always-cheap inference gives $0.426$ and
always-expensive $0.577$. Training the geometry router on
$\Delta_{\mathrm{correct}}$ rather than $\Delta_{\mathrm{prob}}$ recovers most of
its apparent deficit against confidence.}
\label{tab:router}
\end{table}

\begin{figure}[t]
\centering
\includegraphics[width=0.75\textwidth]{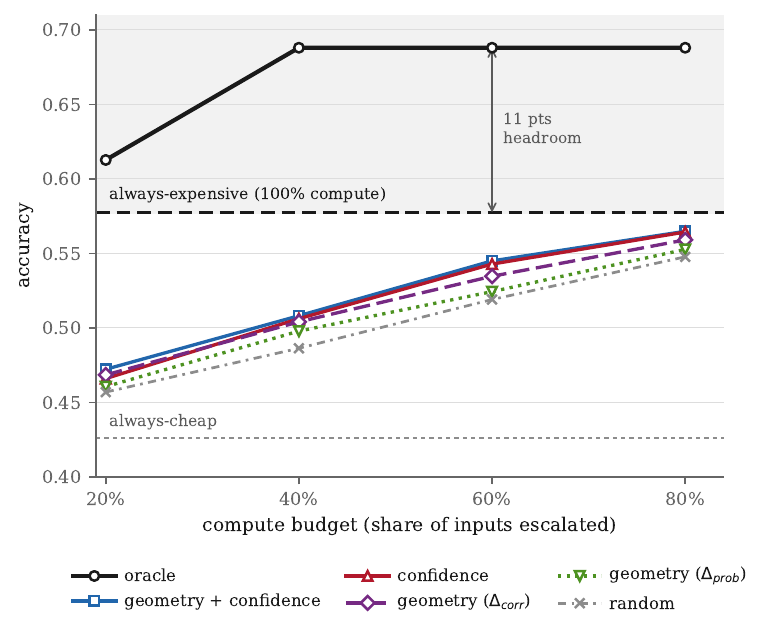}
\caption{Accuracy--compute frontier over the 11 viable ladder languages,
leave-one-language-out; the values are those of Table~\ref{tab:router}. The
shaded band marks accuracy unreachable by any practical method tested: every
routing signal lies below always-expensive inference at every budget, while the
oracle exceeds it by 11 points at 60\% of the compute.}
\label{fig:frontier}
\end{figure}

Table~\ref{tab:router} and Figure~\ref{fig:frontier} report the corrected
experiment. Retraining the geometry
router on the decision-relevant target improves it at every budget---by $+0.007$
at 20\% and $+0.011$ at 60\%---and correspondingly narrows the gap to confidence.
Across ridge penalties $\lambda \in \{1, 10, 100, 1000\}$ the
confidence$-$geometry gap is $-0.0015$ to $-0.0024$ at the 20\% budget, where
confidence wins in 5 of 11 held-out languages, and $+0.0059$ to $+0.0147$ at the
60\% budget, where it wins in 6 to 8. Combining geometry with confidence yields a
further $-0.0005$ to $-0.0062$ over confidence alone, winning in 4 to 6 languages
of 11. The per-language standard deviation is approximately $0.018$ throughout,
larger than every one of these differences.

\claim{Once the routing predictor is trained against the decision-relevant
target, representation geometry and confidence are statistically
indistinguishable across the tested budgets and regularisation settings, and both
remain substantially below the oracle.} Win counts at or near 5.5 of 11 are what
one expects from two signals of comparable value, not from one dominating the
other.

\subsection{Unexploited headroom}
\label{sec:headroom}

The corrected comparison should not be read as a mild success for representation
geometry. A stronger negative result is visible in the same table.
\claim{No practical method tested exceeds always-expensive inference at any
budget} (Figure~\ref{fig:frontier}): at the 60\% budget the best method reaches $0.545$ against $0.577$ for
simply running the expensive model on everything, and at 80\% it reaches $0.565$,
still below. Under these signals, adaptive routing does not pay for itself.

The oracle behaves differently. At the 60\% budget an oracle over
$\Delta_{\mathrm{correct}}$ reaches $0.688$---eleven points above
always-expensive inference while using 40\% less computation.
\claim{The routing problem is therefore genuinely learnable, and neither of the
tested signals captures it.} The gap between the best practical method and the
oracle is approximately 14 accuracy points at the 60\% budget.

We draw the conclusion narrowly. The negative result is not that adaptive routing
is infeasible in this setting; the oracle demonstrates the opposite. Nor is it
that representation geometry is uninformative; \S\ref{sec:swap} establishes a
robust association with one notion of computational benefit. It is that neither
the geometry statistics nor cheap-model confidence extracts enough of the
available benefit to be useful, and that the appearance of a clear baseline win
in our original experiment was substantially an artifact of target misalignment.

\section{Discussion}
\label{sec:discussion}

\claim{Our experiments do not show that representation geometry is useless. They
show that representation structure, statistical association, and decision
usefulness are three different things, and that multilingual evaluation can
conflate them.} Each stage of the investigation separates a pair of these that
would otherwise be run together.

\paragraph{Evaluation validity.} The contamination result does not establish that
AfriXNLI is invalid, and we do not claim it. A translated benchmark that
deliberately preserves its source-language subsets is behaving as documented; the
difficulty arises from the pairing of such a benchmark with checkpoints trained on
the source. The narrower conclusion we draw is that
\claim{model--benchmark pairings require contamination-aware reporting}: a result
on AfriXNLI English, French, or Swahili obtained with an XNLI-trained checkpoint
is not interpretable as a measure of capability, and the dev--test gap of
Table~\ref{tab:devtest} offers a cheap diagnostic for detecting the problem
without access to training data. This matters most where it is least likely to be
checked---the high-resource configurations of a low-resource benchmark, which are
typically used as controls rather than as objects of study.

\paragraph{Model selection.} \S\ref{sec:f2} implies that a cascade cannot be
constructed by sorting checkpoints by parameter count. The relevant finding is
not that the larger model is globally inferior---it is not, and the aggregate
difference is not distinguishable from zero---but that the ordering is
language-dependent, with differences exceeding ten accuracy points in six of
fifteen languages and running in both directions. For a practitioner building a
multilingual cascade this converts an assumption into a measurement: which
checkpoint is the ``expensive'' one must be determined per language, and in some
languages the answer inverts.

\paragraph{Representation confounding.} We regard
\S\ref{sec:f3a} as the most generalisable result in the paper, because it
replicates across three representation spaces produced in different ways: no two
share a training objective, and only two of the three share a pre-training
corpus. Angular dispersion
is consistently far more language-determined than effective rank, and a statistic
that is roughly half between-language variance cannot support a pooled
correlation that is interpreted as an example-level relationship. The practical
form of this observation is a reporting convention rather than a prohibition:
\claim{report $\eta^2$ alongside any pooled multilingual correlation}, so that
readers can see how much of the reported relationship could be carried by
language identity alone.

We are equally explicit about the boundary. We hypothesised a stronger,
quantitative version---that $\eta^2$ predicts the magnitude of pooling bias
across features---and it did not survive the addition of a third source. We
therefore establish an ordering of statistics by language dependence and a
mechanism by which pooling misleads, but not a law relating the two.

\paragraph{What ``benefit from computation'' means.} The result we would
emphasise most is \S\ref{sec:swap}, because it explains the others rather
than merely adding to them. Effective rank is associated with
$\Delta_{\mathrm{prob}}$ and not with $\Delta_{\mathrm{correct}}$; cheap-model
confidence is associated with $\Delta_{\mathrm{correct}}$ and not with
$\Delta_{\mathrm{prob}}$. Both quantities are reasonable formalisations of ``this
example benefits from more computation,'' and they correlate at only $+0.655$.
\claim{A feature can therefore be genuinely informative about one
operationalisation of computational benefit while being irrelevant to---or, for
layer-12 angular dispersion, oppositely related to---the objective a system is
actually evaluated against.}

This is what makes the failure mode difficult to notice. A researcher who
measures a representation statistic against a plausible difficulty target, finds
a robust and highly significant association, and concludes that the statistic
captures difficulty will have done nothing methodologically unusual. Our own
initial routing experiment made precisely this error, and the error was invisible
in the routing results themselves; it became visible only when the two targets
were placed side by side.

\paragraph{Practical routing.} The conclusion we draw from \S\ref{sec:f4} is
not that adaptive inference does not work. The oracle reaches $0.688$ at the 60\%
budget against $0.577$ for always-expensive inference, so under the tested models
and languages there is a substantial and genuinely learnable allocation problem.
\claim{What the evidence supports is that the tested representation and
confidence signals do not recover enough of that headroom to make routing
worthwhile under the tested models, signals, and compute budgets}---no practical
method we evaluated exceeds simply running the expensive model on every input.
The gap between what is achievable and what these signals achieve is roughly 14
accuracy points, and closing it is an open problem rather than a closed one.

\paragraph{Recommendations.} Four practices follow directly from the failures
above. Audit translated benchmarks for lineage, and report the dev--test gap when
evaluating with checkpoints whose training data overlaps the source. Do not order
cascade rungs by parameter count without per-language verification. Report
$\eta^2$ alongside pooled multilingual correlations. And
\claim{state which notion of computational benefit a difficulty signal is
validated against, and train routing predictors on the quantity the evaluation
scores.}

\paragraph{Future work.} Three directions follow from what the experiments
exposed rather than from speculation. First, contamination-controlled
multilingual benchmarks, or at minimum published lineage metadata sufficient for
practitioners to exclude affected configurations automatically. Second,
task-specific and language-specific capacity ladders, constructed by measurement
rather than by parameter count, which would also supply the wider compute range
our frozen encoder pair could not. Third, routing targets defined directly from
the deployment objective, with representation signals validated against that
objective rather than against a proxy. A seed-averaged marginal-gain target is a
natural next experiment in this last direction, and one our frozen setup could
not provide (\S\ref{sec:limitations}).

\section{Limitations}
\label{sec:limitations}

\paragraph{The marginal-gain targets.} The most significant limitation concerns
how the targets of Equations~\ref{eq:dprob} and \ref{eq:dcorr} were constructed,
and it has two parts.

The first is a matter of interpretation. $\Delta_{\mathrm{prob}}$ is a
well-defined and useful quantity for the descriptive analysis of
\S\ref{sec:f3b}: it measures continuously how much escalation changes the model's
probability on the correct label, and a continuous target admits far more
statistical power than a ternary one. \claim{But our own routing experiment shows
that $\Delta_{\mathrm{prob}} \neq \Delta_{\mathrm{correct}}$, and
$\Delta_{\mathrm{prob}}$ should therefore not be treated as a universal measure of
computational benefit.} It is one operationalisation among several, and
\S\ref{sec:swap} demonstrates that the choice determines which signals appear
informative.

The second is more restrictive. \claim{Because the models were frozen and no
repeated fine-tuning runs were available, the study could not construct the
initially planned seed-averaged marginal-gain target. The stability of the
measured $\Delta_{\mathrm{prob}}$ relationship therefore reflects variation across
examples and languages within fixed checkpoints, but does not establish how
stable the relationship would be across independently trained model instances.}
Part of the variance in Equation~\ref{eq:dprob} is checkpoint idiosyncrasy rather
than example difficulty, and we cannot separate the two. This bears on
\S\ref{sec:f3b} and \S\ref{sec:f4}; per Table~\ref{tab:dependency}, it does not
bear on \S\ref{sec:f1}--\S\ref{sec:f3a}.

\paragraph{Benchmark lineage.} Excluding \texttt{eng}, \texttt{fra}, and
\texttt{swa} removes the direct surface-overlap problem, but the 15 remaining
configurations are translations of XNLI instances that the checkpoints
encountered in other languages. Cross-lingual contamination cannot be ruled out,
and we have no uncontaminated African-language NLI benchmark of comparable
coverage against which to estimate its size. Our temperature calibration, fitted
on the African development splits, shares this lineage; the fitted quantity is a
single scalar per model and all reported results are on test, but the pipeline as
a whole is not contamination-free.

\paragraph{Checkpoint selection.} We examine three particular publicly available
NLI checkpoints, not the space of multilingual architectures. The
capability-ordering result therefore supports language-dependent capability
\emph{for these models}, and should not be read as a general law about parameter
count. All three are also XNLI-derived, which makes them a correlated sample
rather than three independent observations.

\paragraph{Representation statistics.} The three-source replication is stronger
than a two-model analysis, but three encoders of similar depth remain a limited
sample of model families, and effective rank, spectral concentration, and angular
dispersion are a limited sample of possible geometric descriptors. The statistics
are additionally computed over the token dimension of a single sequence, which
makes them partly length-determined by construction; we control for length
statistically rather than by design.

\paragraph{Single task and coverage.} Natural language inference is the entire
experimental task. We do not claim that the target-alignment or routing
conclusions transfer automatically to generation, question answering,
classification, or agentic reasoning. Within the task, the routing experiments
cover only the 11 of 15 languages in which the cheap rung exceeds chance, and
only the single MiniLM$\rightarrow$mDeBERTa model pair.

\paragraph{Post-hoc analyses.} Two analyses were not part of the initial design
and are identified as post-hoc throughout: the addition of XLM-R-base as a third
representation source (\S\ref{sec:f3a}), which we ran to test a generalisation
that subsequently failed, and the dual-target analysis (\S\ref{sec:swap}), which
we ran after identifying the routing-target mismatch. Both are explanatory rather
than confirmatory, and neither has been replicated on independent data.

\paragraph{The oracle is not deployable.} The oracle in Table~\ref{tab:router}
has access to the outcome it is asked to predict. \claim{It establishes an upper
bound on what perfect per-example allocation could achieve under this cascade,
not evidence that a practical system can approach that level.} The 14-point gap
we report should be read as the size of the opportunity, not as a deficit
attributable to any particular method.

\claim{Taken together, this paper identifies several measurement and evaluation
barriers to finding useful adaptive signals; it does not establish what the
optimal adaptive signal is.}

\section{Conclusion}
\label{sec:conclusion}

We set out to determine whether internal representation statistics can estimate,
per example, the value of escalating an input to a more capable model in
low-resource multilingual NLI. Answering that question required first
establishing what our evaluation measured. AfriXNLI shares 1{,}047 of its 1{,}050
English examples verbatim with XNLI, and the checkpoints commonly used to
evaluate on it are trained on XNLI, so its English, French and Swahili
configurations cannot serve as clean evaluations or as high-resource controls.
Among the remaining 15 languages, parameter count does not order capability: the
larger of our two main checkpoints is better in seven languages and worse in
eight, with no significant aggregate difference. Across three multilingual
representation spaces, angular dispersion is consistently dominated by language
identity and effective rank consistently is not, so pooled correlations inflate
the former and mask the latter. And the association that survives every control
is specific to the target it was measured against: effective rank predicts the
probability gain from escalation but not whether escalation changes the
prediction, while cheap-model confidence does the reverse.

\claim{Under the tested models, signals, and compute budgets, the evaluated
signals did not recover enough of the available headroom to make adaptive routing
preferable to always-expensive inference.} We do not conclude that adaptive
inference is infeasible here; an oracle reaches $0.688$ at the 60\% budget
against $0.577$ for always-expensive inference, so a substantial and learnable
allocation problem exists. What we conclude is narrower and, we think, more
useful: \claim{representation structure, statistical association, and decision
usefulness are three distinct properties, and multilingual evaluation can
conflate all three.}

\section*{Reproducibility}
Code, configurations, cached model outputs and a results ledger recording every
reported number with its sample, statistic, uncertainty and limitation are
available at \url{https://github.com/qeinstein/adaptive-computation}, at the tag
\texttt{paper-v1}. No model was
fine-tuned. All inference was performed once on a laptop without a discrete GPU and
cached; every analysis in this paper, including all bootstrap and permutation
procedures, re-runs from those caches in minutes and requires no accelerator. The
ledger also records the analyses that were superseded during the study, including
the initial misaligned routing experiment of \S\ref{sec:targetalign}.

\section*{Ethics and Data Statement}
This work uses only publicly released benchmarks and publicly released model
checkpoints, and involves no human subjects and no new data collection. Our
contamination findings concern the interaction between a benchmark's construction
and a checkpoint's training data; both were documented by their authors, and we
intend the analysis as a caution about how such artefacts are combined in
evaluation rather than as criticism of anyone who released them. We note that
undetected contamination is likely to distort measured progress on low-resource
languages specifically, since the affected configurations are typically the
high-resource controls against which low-resource results are compared.

\appendix

\section{Per-Language Capability Differences}
\label{app:perlang}

\begin{table}[h]
\centering
\small
\begin{tabular}{lrlrlr}
\toprule
Lang & $\delta$ & Lang & $\delta$ & Lang & $\delta$ \\
\midrule
\texttt{ibo} & $+0.220$ & \texttt{lug} & $+0.033$ & \texttt{zul} & $-0.003$ \\
\texttt{sna} & $+0.200$ & \texttt{twi} & $+0.015$ & \texttt{yor} & $-0.010$ \\
\texttt{sot} & $+0.197$ & \texttt{lin} & $+0.008$ & \texttt{ewe} & $-0.037$ \\
\texttt{kin} & $+0.147$ & \texttt{wol} & $-0.002$ & \texttt{hau} & $-0.080$ \\
             &          &              &          & \texttt{xho} & $-0.087$ \\
             &          &              &          & \texttt{orm} & $-0.127$ \\
             &          &              &          & \texttt{amh} & $-0.145$ \\
\bottomrule
\end{tabular}
\caption{Accuracy difference $\delta=$ mDeBERTa $-$ XLM-R-large per language.}
\label{tab:perlang}
\end{table}

\section{Stability of the Effective-Rank Association}
\label{app:stability}

\begin{table}[h]
\centering
\small
\begin{tabular}{lr}
\toprule
Stability test & Result \\
\midrule
Bootstrap, 300$\times$ within-language resample & $-0.127$, 95\% $[-0.142,-0.111]$ \\
Subsample 25\%, 40 seeds & $-0.122 \pm 0.019$ \\
Subsample 50\%, 40 seeds & $-0.129 \pm 0.011$ \\
Leave-one-language-out & $[-0.138, -0.117]$ \\
Label permutation, 20$\times$ & $-0.001$, $\max|\rho| = 0.026$ \\
\bottomrule
\end{tabular}
\caption{All resampling variants remain negative; the permutation null is
centred at zero with maximum magnitude five times smaller than the effect.}
\label{tab:stability}
\end{table}

\end{document}